# Hyperparameter Optimization and Reproducibility in Deep Learning Model Training

Usman Afzaal[1]*, Ziyu Su[1], Usama Sajjad[1], Hao Lu[2], Mostafa Rezapour[3], Metin Nafi Gurcan[2], Muhammad Khalid Khan Niazi[1]

[1]Department of Pathology, College of Medicine, The Ohio State University Wexner Medical Center, Columbus, OH, USA
[2]Center for Artificial Intelligence Research, Wake Forest University School of Medicine, Winston-Salem, North Carolina, USA
[3]Wake Forest Institute for Regenerative Medicine, Wake Forest University School of Medicine, Winston-Salem, North Carolina, USA
***Corresponding author**: Usman Afzaal (usman.afzaal@osumc.edu)

**Abstract**

Reproducibility remains a critical challenge in foundation model training for histopathology, often hindered by software randomness, hardware non-determinism, and inconsistent hyperparameter reporting. To investigate these issues, we trained a CLIP model on the QUILT-1M dataset and systematically evaluated the impact of different hyperparameter settings and augmentation strategies across three downstream histopathology datasets (PatchCamelyon, LC25000–Lung, and LC25000–Colon). Despite variability across runs, we identified clear trends: RandomResizedCrop values of 0.7-0.8 outperformed more aggressive (0.6) or conservative (0.9) settings, distributed training without local loss improved stability, and learning rates below 5.0e-5 consistently degraded performance across all datasets. The LC25000 (Colon) dataset consistently provided the most reproducible benchmark. These findings highlight that reproducibility in computational pathology depends not only on transparent documentation but also on carefully chosen experimental configurations, and we provide practical rules to guide future efforts in developing reproducible foundation models for digital pathology.

# 1 Introduction

## 1. 1 The Issue of Reproducibility

Concerns about reproducibility in science have been growing, with widespread agreement on an ongoing crisis[1]. Despite the deterministic nature of computers, reproducibility remains a concern in computer science experiments[2], and machine learning in no exception. The ICLR 2018 Reproducibility Challenge confirmed similar concerns among experts[3,4]. It is vital to ensure that Deep Learning (DL) models are reproducible for a variety of reasons, including but not limited to, future training of the algorithms[5], testing[6], model debugging[7], legal compliance[8] and facilitating the research community in open science[9,10]. However, before examining these challenges in detail, it is important to clarify what reproducibility means in the context of deep learning, as the field lacks consensus on precise definitions.

In DL, reproducibility generally means replicating an experiment and achieving the same metrics. However, no single definition exists, and multiple interpretations persist. Drummond defines replication as exact repetition and reproducibility as obtaining similar results with different experiments[11]. Stodden describes replication as re-running with provided code and data, while reproduction extends to regenerating findings with partial independence from original code/data[12]. Goodman et al.[13] propose three terms: (1) methods reproducibility, following the same procedures; (2) results reproducibility, obtaining the same results in an independent but closely matched study; and (3) inferential reproducibility, drawing the same conclusions from independent work. Gundersen and Kjensmo outline reproducibility levels in AI based on documentation[14], while Tatman et al. define low, medium, and high levels depending on whether paper, code, data, and environment are shared[4,15]. Plesser provides an overview of the different definitions for reproducibility[16,17]. For DL training, a model is reproducible if the same setup (dataset, code, hyperparameters, hardware, and software) produces comparable results[7,18]. In practice, this often involves comparing distributions from multiple training runs, expecting significant overlap and no statistical differences. Despite these various frameworks for understanding reproducibility, achieving it in practice remains a significant challenge across the deep learning community.

Nevertheless, reproducibility remains difficult[5,7,15,17,19-21]. Without it, claims of improvement are unsubstantiated, debugging becomes impractical[22], and cross-lab comparisons unreliable. Researchers often compare algorithms to reported results without re-running them, risking misleading conclusions. Irreproducible models also waste resources[23] and discourage newcomers. To understand why reproducibility is so difficult to achieve, we must examine the specific sources of variability that occur within deep learning experiments.

**1. 2 Causes of Variability**

DL model variability arises from (a) poor documentation and lack of guidelines[14], (b) software environment differences, and (c) hardware non-determinism[24]. Each can shift model weights enough to alter conclusions [4]. Lack of code/data sharing also contributes to this problem[25]. Unlike deterministic programs, DL involves stochastic optimization, and as such, their output might not be reproducible and trustworthy in some instances[26]. This is particularly true for complex deep learning networks, which require trillions of calculations per experiment[27]. As a direct consequence of the aforementioned reasons, even if independent researchers are able to execute the provided instructions and code, the results of the algorithm are not guaranteed to be statistically similar, i.e., *p-value* for the relevant test is lower than 0.05.

**1. 2. 1 Randomness within Software Environments** Randomness plays a crucial role in deep learning model training, with aspects such as batch ordering, data shuffling, and weight initialization being key to building robust and high-performing DL models[7,28]. However, such randomness prevents the DL models from being reproduced, as different starting points can lead to divergent solution[29]. For instance, in Stochastic Gradient Descent (SGD), a random subset of data (known as a mini batch) is chosen in each iteration to compute the gradient. This random selection of mini-batches can influence the trajectory of the optimization process[30,31]. Framework and library versions (like TensorFlow[32] or PyTorch[33], CUDA, MKL) which are used for numerical computation, can also cause variations.

**1. 2. 2 Non-Determinism in Hardware Components** Hardware differences are a major source of non-determinism. Hong et al. found output variance from hardware/compiler differences comparable to that from altered initial conditions[34]. GPUs, essential for large models[19], introduce rounding errors in parallel floating-point operations[7,19,35]. Furthermore, different GPU architectures might handle parallel computations differently and yield distinct results[36]. NVIDIA's CUDA Deep Neural Network library (cuDNN)[37], which is widely used in DL, accelerates training but introduces non-deterministic algorithms[22,36,38]. Auto-tuning, where GPU-specific libraries automatically select the optimal primitive operations, further contributes to variability[7,19,28].

**1. 2. 3 Incomplete Information** The primary factor behind many reproducibility issues, if not all, is the absence of complete information. Occasionally, this gap is deliberate, such as when data cannot be shared due to privacy constraints[23]. Various frameworks and checklists[19,39-42] recommend recording model design, hyperparameters, dataset versions, preprocessing steps, and software versions. Documenting hardware and software details, outlining preprocessing steps, elucidating model architecture specifics, and listing external dependencies are additional vital measures for reproducibility[43-53]. Providing instructions or scripts for experiment replication, coupled with clear articulation of licensing terms and ethical considerations, ensures that researchers can readily comprehend, replicate, and extend foundational models. Unfortunately, even after taking these considerations into account, we have encountered reproducibility issues while training DL foundation models.

To better understand how reproducibility challenges manifest in real-world deep learning applications, particularly in computational pathology, we turn to a case study involving contrastive learning. This paradigm has gained prominence for its scalability and effectiveness in pretraining large models, yet its reproducibility in domain-specific contexts like histopathology remains underexplored. By examining the CLIP model trained on Quilt-1M, we aim to investigate how foundational models behave under the constraints and variability discussed above, and whether current practices are sufficient to ensure reproducibility in high-stakes medical applications.

**1. 3 Case Study**

Contrastive learning has emerged as a powerful pretraining method, aligning embeddings of similar (positive) pairs while separating dissimilar (negative) pairs[54-58]. Contrastive learning has demonstrated its high effectiveness and scalability in the context of pretraining dual-encoder image-text models[59,60]. These models have shown exceptional performance across various downstream visual recognition tasks. The motivation behind such models is leveraging the broader information available from raw text about images, as opposed to simple categorical labels. This training paradigm enables the network to learn associations between images and text and to align these visual and textual representations within a shared latent space. It does so by a simple pre-training task of predicting which caption goes with which image is an efficient and scalable way[59]. Similar methods which learn directly from raw have revolutionized natural language processing (NLP) over the past few years[61]. Likewise in the field of computer vision, research has showcased that with extensive pretraining while utilizing vast datasets obtained from the web, comprising noisy image-text pairs, can achieve the dual objectives of learning well-aligned representation spaces between images and language[59]. Furthermore, these aligned latent spaces can be effectively transferred to perform downstream tasks, including image classification, zero-shot classification[59,60,62-64]. Given these promising capabilities of contrastive learning in histopathology, we examine the reproducibility challenges that arise when training such models and propose systematic approaches to address them.

This paper discusses the intricate problems we encountered while building our machine learning model in terms of reproducibility and subsequently provide heuristics to tackle these problems. Our approach entails a record-and-replay based technique for mitigating non-determinism arising from the software and the hardware, included in which is a study of the hyperparameters used for training the CLIP model and their observed effects on the network training, along with discovered trends. Among other findings, our results suggest the importance of data augmentation for CLIP model training. We also go ahead and propose a systematic framework to work out the issue of reproducibility in deep learning based on our experience.

In summary, our main contributions are:

- We investigate the impact of training hyperparameters and augmentation strategies on the CLIP contrastive learning approach.

- We present a methodological framework for documenting hyperparameters, designed to simplify the process of reproducing similar results for future studies.

**2 Related work and Background**

**2. 1 Large scale visual language pretraining**

Contrastive learning has proven to be a powerful and scalable approach for pretraining dual-encoder models that integrate image and text. These models have demonstrated proficiency across various visual recognition tasks, particularly in the medical imaging domain. Seminal projects like CLIP[59] and ALIGN[60] have shown that models pretrained on vast and varied datasets of image-caption pairs from the web can achieve robust zero-shot learning. They do this by harnessing prompts that tap into the image-text alignment established during pretraining. In the domain of histopathology[62,65], researchers have utilized thousands of histopathology image-caption pairs to enhance the efficiency and zero-shot learning abilities of these models.

**2. 2 Reproducibility efforts**

Olorisade et al.[66] analyzes the reproducibility of text mining studies in systematic reviews, identifying key factors affecting reproducibility and evaluating 30 studies for such information, ultimately providing recommendations to enhance reproducibility in research reporting. Renard et al.[67] conducts a literature review on deep learning in medical image segmentation, discussing its variability and reproducibility challenges, and proposes three key recommendations for addressing these issues, including a framework description, analysis of variability sources, and an effective evaluation system for segmentation results. Sugimura et al.[23] addresses the challenges of model reproducibility in machine learning, discussing encountered problems and introducing a framework with four main components (data, feature, scoring, and evaluation layers) to ensure precise replication and transformation reusability, thereby enhancing both the speed and reliability of model experimentation. Ahmed et al.[68] studied reproducibility in machine learning, focusing on the challenges posed by pseudo-random numbers in model generation. The authors presented examples of randomness impact and investigated methods for controlling random number generation to enhance the reproducibility and trustworthiness of ML models. Chen, B. et al. [19] studied reproducibility issues in deep learning, particularly focusing on the challenges posed by software randomness and hardware non-determinism. A three-part approach for training DL models was proposed, (1) a set of criteria for evaluating reproducibility across different domains, (2) a framework that combines record-and-replay and profile-and-patch techniques to manage software and hardware variability, and (3) a guideline for executing a reproducible DL training process. Wagner et al.[69] investigated the impact of hyperparameters and data augmentation strategies on Self-Supervised Learning (SSL) performance, using Bayesian optimization to enhance the SimSiam SSL approach. GroupAugment, a novel automated data augmentation algorithm optimized for SSL was introduced, demonstrating its effectiveness across various datasets and highlighting the critical, yet often overlooked, role of data augmentation in SSL. Building on these prior efforts, we develop a comprehensive methodological framework that addresses reproducibility challenges specific to histopathology foundation model training.

## 3 Methods and Materials

### 3. 1 Training Overview

Histopathology whole-slide images are rich in information, with each image patch containing unique, intricate patterns essential for tissue analysis. Reducing this vast information to a single label oversimplifies the complexity inherent in histopathology, a field encompassing thousands of evolving disease subtypes[70]. It underscores the importance of developing more expressive and interconnected representations that extend beyond the capabilities of a single categorical label. Consequently, natural language descriptions emerge as a powerful tool, offering a comprehensive perspective by connecting the diverse features of sub-patch structures in histopathology[71,72].

Previously, there have been successful examples of task-specific model development in histopathology, where models are trained separately for each task, like lymph node metastasis detection[73-76]. Nevertheless, this approach becomes impractical when attempting to scale it to encompass the hundreds of tumor types spread across numerous organ sites, considering many tumor types are either inadequately represented in publicly available datasets or lack sufficient samples for comprehensive model development [62].

As a solution, a dual-encoder CLIP model can be pre-trained in a task-agnostic fashion on the web-sourced, large-scale Quilt-1M image-caption pairs pathology dataset. CLIP acquires knowledge directly from raw text linked to images, offering a richer context value than relying solely on class labels. Therefore, we use a large-scale image-caption pathology dataset QUILT-1M to finetune a vision-language CLIP model using a contrastive objective between the two modalities. CLIP is pre-trained using a contrastive objective, which, in this case, focuses on learning to distinguish between similar (positive) and dissimilar (negative) pairs of data points and is primarily used to train unsupervised and self-supervised learning models, ultimately leading to the vision-text embedding. Consequently, this pre-trained representation space can then be applied to perform further downstream tasks like zero-shot image classification, where the encoders are then "prompted" with text from the label space.

We evaluate our approach on 3 external histopathology datasets across different sub-pathology domains, reporting zero-shot classification results and comparing against baseline CLIP models trained on QUILT-1M. For our model, we explore the influence of different training hyperparameters and augmentation strategies. While

CLIP and similar models have seen widespread applications in natural image processing[68-71] and specialized medical fields like radiology[77-81] and pathology[62,82], there is limited exploration of how different training settings and augmentation methods affect their performance. To address these lesser-studied aspects, we conduct an extensive search over various configurations to fine-tune the CLIP model on the Quilt-1M dataset. Our research is two-fold: we investigate the influence of training hyperparameters and examine the role of data augmentation techniques. Furthermore, we present ablation studies with regard to the effects of multiple hyperparameters and the observed trends.

### 3. 2 Training dataset

For this work, we have used the publicly available dataset Quilt-1M[65], curated mostly using YouTube videos. Initially, the dataset QUILT was collected from valuable educational, 4504 histopathology narrative videos from expert pathologists on YouTube spanning over 1087 hours with 802,148 image-text pairs. A mixture of models including, large language models, handcrafted algorithms, human knowledge databases, and automatic speech recognition were used to align image and text pairs from the videos. The mean length of the associated text captions in QUILT is 22.76 words. The images contained in the dataset are of varying magnification scales (0-10x, 10-20x, 20-40x), with (280K, 75K, 107K) images for each scale, respectively.

QUILT does not overlap with any current open-access histopathology data sources. And so, this allows the dataset to be further been merged with other open-source datasets available online to create an even larger dataset, namely QUILT-1M. The sources for histopathology image-text open-access pairs include: (a) LAION[83], (b) X (formally known as Twitter Inc.), and (c) PubMed. QUILT-1M contains roughly a million, pathology related image-text pairs, which makes it the largest public vision-language histopathology dataset to date.

### a. PubMed Open Access Articles:

PubMed open-access from 2010-2022 was accessed with keywords (pathology, histopathology, whole-slide image, H&E, and 148 keywords from a histopathology glossary[84], extracting 62,458 histopathology image-text pairs, from 109,518 unique histopathology articles. The average caption length for this dataset is 54.02 tokens.

### b. LAION:

The Large-scale Artificial Intelligence Open Network (LAION-5B)[83] curated over 5 billion image-text pairs from across the internet. This collection also includes a substantial amount of histopathology-related data. From this, 22,682 image and text pairs were retrieved to merge with the QUILT dataset.

### c. Twitter Data from OpenPath:

A list of tweets curated by Huang et al.[72] was utilized, which contained a total of 55,000 unique tweets and 133,511 unique image-text pairs. A one-to-many relationship was maintained where many images were matched with multiple captions.

### 3. 3 Model architecture

CLIP is a dual-encoder architecture with an image and a text encoder. A transformer is used as the text encoder, as for the image encoder a version of ResNet[85] or Vision Transformer (ViT)[86] is used. During training, CLIP receives a batch containing N image and text pairs as inputs for their respective encoders. It is then trained to predict which of the N x N possible pairs in the batch are connected. In each batch, the ith image corresponds to the ith textual caption. CLIP learns a multi-modal embedding space by jointly training an image encoder $f(.;\theta)$ and text encoder $g(.;\Phi)$, to maximize the cosine similarity of the aligned image and text embeddings of the N real pairs in the batch while minimizing the cosine similarity of the embeddings of the $N^2$-N incorrect pairings. This is achieved by employing a cross-model contrastive loss, formulated as a temperature scaled N-way classification[87]. For a given batch of $N$ image and text pairs, denoted as $\{(x_n, t_n)\}_{n=1,\dots,N}$, which are fed into their respective encoders, a learnable linear projection is applied. This projection transforms each encoder's output into a multi-modal embedding space of the same dimension. $l_2$-normalized visual and text embeddings are evaluated from these inputs, respectively as $u_n = \frac{f(x_n; \theta)}{\|f(x_n; \theta)\|}$ and $v_n = \frac{g(t_n; \theta)}{\|g(t_n; \theta)\|}$. The cosine similarity is then calculated between the $l_2$-normalized embeddings $u$ and $v$. The cosine similarity is calculated between a fixed image embedding $u_i$ and every text embedding $v_j$ in the batch as $sim(u_i, v_j) = u_i^T v_j$. The same process is symmetrically applied for a fixed text embedding $v_j$ across all image embeddings $u_i$. The model then employs bi-directional contrastive learning, operating in two directions: image-to-text ($i \to t$) and text-to-image ($t \to i$). During this process, a symmetric cross-entropy loss, known as the InfoNCE loss, is applied to the similarity scores in both directions, ensuring the model captures the intricate relationships between images and texts[59].

$$L_{i \to t} = \sum_{i=1}^{N} \log \left( \frac{\exp(\tau sim(u_i, v_i))}{\sum_{j=1}^{N} \exp(\tau sim(u_i, v_j))} \right)$$

$$L_{t \to i} = \sum_{j=1}^{N} \log \left( \frac{\exp(\tau sim(v_j, u_j))}{\sum_{i=1}^{N} \exp(\tau sim(v_j, u_i))} \right)$$

$$L = -\frac{1}{2N} (L_{i \to t} + L_{t \to i})$$

In this setup, $u_i$ and $v_i$ represent the $l_2$-normalized embeddings for the $i$-th image and its corresponding text. The temperature parameter $\tau$ regulates the logits' range in the SoftMax function. With a batch of $N$ image-text pairs, the task becomes an $N$-way classification problem, where the correct text (or image) for each image (or text) is treated as the 'target', and the other $N-1$ indices in the batch, representing mismatched texts (or images), serve as 'negatives'. The contrastive loss employs temperature-scaled cosine similarity scores between the embeddings as logits, with the objective of minimizing the cross-entropy loss to clearly distinguish 'target' from 'negative' pairs in the model's learning process[62].

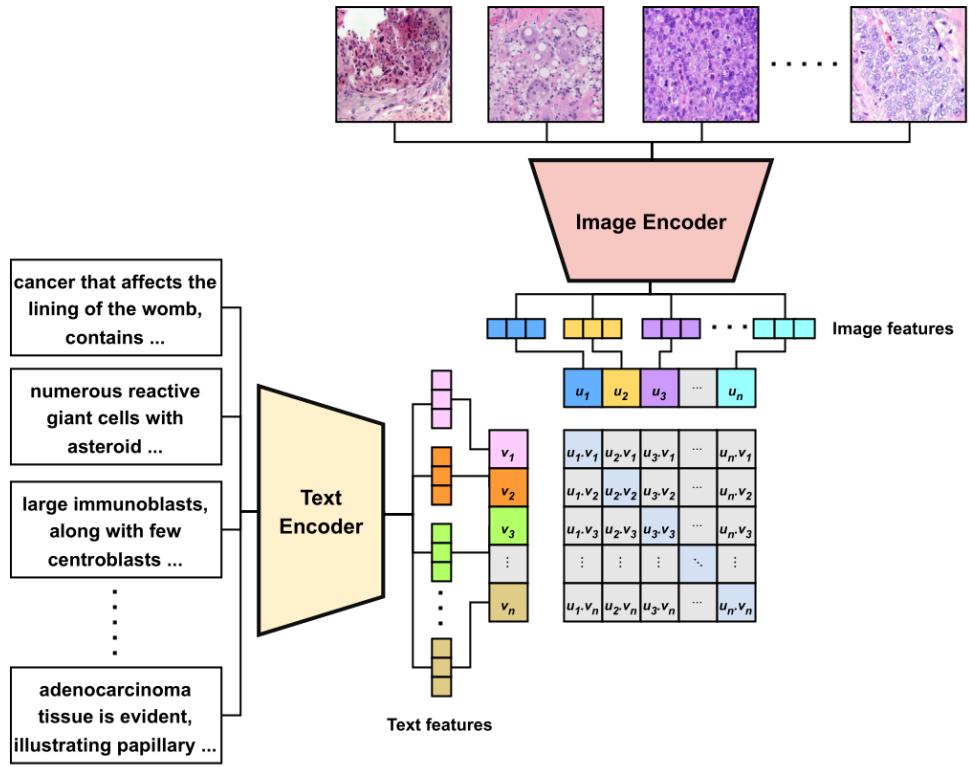

Figure 1 Summary of the CLIP architecture. CLIP jointly trains an image and a text encoder to learn a multi-modal embedding space by maximizing the cosine similarity of the correct pairs in a batch of (image, text) training examples.

In our model, we initialize the image and text encoders from pretrained OpenAI weights[59], before finetuning it on the QUILT-1M dataset. CLIP was originally pre-trained on a dataset of 400 million internet-sourced images and their corresponding text captions. This extensive pre-training enables it to perform comparably to models trained on benchmark datasets for most tasks, despite not using those benchmark datasets nor undergoing additional training. The architecture of the CLIP model is shown in Figure 1. For the image encoder, CLIP closely follows the implementations in [86] with some minor modifications of adding an additional layer normalization to the combined patch and position embeddings before the transformer block, while also utilizing a different initialization method. We leverage ViT-B/32 as the image encoder, where "B" signifies the "Base" variant with a patch size of 32. For encoding text, we adopt GPT-2[88], which supports a context length of up to 77 words. The text encoder operates on a lower-cased byte pair encoding (BPE) of the text[89]. Having described the model architecture and training objective, we now outline the specific implementation details and reproducibility measures employed in our experiments.

### 3. 4 Implementation Outline

In this research, all model implementations are based on the OpenCLIP open-source repository[90,91]. The experiments were conducted using PyTorch and utilized 4 NVIDIA A100 GPUs. A virtual environment using Anaconda (anaconda.org) was created for the model training. To ensure consistency, we provide a detailed $requirements.txt$ file to set up the virtual environment with the necessary libraries and their specific versions. This file is accompanied by template code for training the models, aimed at eliminating any discrepancies.

A significant challenge we encountered was the inability to replicate experiments consistently. Despite experimenting with various hyperparameter ranges and model configurations, we could not reproduce identical results for the same settings. This issue was primarily due to the non-determinism inherent in the GPUs. To address this, we fixed the seed value for several components on each GPU, including (a) the PyTorch random number generator, (b) the Numpy random number generator, (c) the Python random number generator, (d) the CUDA random number generator, and (e) the seed for all GPUs. Furthermore, we disabled the CuDNN auto-tuning feature, which, while optimizing performance, can introduce non-deterministic behavior. For reproducibility, it is also essential to force CuDNN to use deterministic algorithms, especially in GPU use, as some CuDNN operations are not inherently deterministic by default. The pseudocode for setting up the random seeds in a multi-GPU environment is presented in Figure 2.

```
def random_seed(seed=0, rank=0):
    torch.manual_seed(seed + rank)
    np.random.seed(seed + rank)
    random.seed(seed + rank)
    torch.cuda.manual_seed(seed) # use if you are using CUDA (NVIDIA GPUs)
    torch.cuda.manual_seed_all(seed) # use in case you are using multi-GPU
    torch.backends.cudnn.benchmark = False # CuDNN auto-tuning
    torch.backends.cudnn.deterministic = True
```

Figure 2 Pseudocode to properly initialize the random seeds in a multi-GPU environment.

The rank argument refers to process rank within the distributed training. It ensures that different processes, even if running the same code with the same global seed, will generate different sequences of random numbers. In our case, we set the same seed for the CUDA operations in PyTorch for all GPUs as shown in `torch.cuda.manual_seed(seed)`. However, the results are reproducible in subsequent runs even after adding to it the *rank* argument.

### 3. 5 Implementation details

For all experiments, we have compiled common hyperparameters in Table 1. The initial settings for most hyperparameters were informed by similar prior studies[59,62,65]. Each model was trained using images of size 512, before randomly cropping them to the desired size of 224. For this purpose, we utilized the OpenAI CLIP pretrained network as our starting point. Our approach involved extensive hyperparameter tuning, wherein we experimented with various settings for the learning rate, scheduler, number of epochs, augmentation techniques, weight decay, warmup steps, and other model configurations. This was done in an effort to find the optimal combination of parameters to enhance the model's performance.

Table 1 Shared hyperparameters across all models in the training process.

| Hyperparameter | Value |
| --- | --- |
| Seed | 0 |
| Batch size (per gpu) | 1024 |
| Pre-training | CLIP [59] |
| Optimizer momentum | B1, B2 = 0.9, 0.98 |
| eps | 1.0e-6 |
| Image mean | (0.48145466, 0.4578275, 0.40821073) |
| Image std | (0.26862954, 0.26130258, 0.27577711) |
| Optimizer | AdamW [Adw] |

The configurations tested include: (a) Gathering the features, which involves collecting image features from all processes and concatenating them into a single tensor. This is a common practice in distributed training, where the model operates across multiple GPUs or machines, and there's a need to aggregate the outcomes from all devices into one tensor; (b) The precision of floating-point calculations; and (c) Local Loss calculation, which entails computing the loss using local features in conjunction with global ones (as opposed to fully realizing a global-to-global matrix). In this context, Local refers to the features present on the current GPU.

Table 2 (Set 1) Hyperparameter selection for base model configuration. "Gather with grad" here refers to gathering the image features from all the processes.

| Hyperparameters/ Experiments | Warmup (steps) | Peak learning rate | Scheduler | Weight decay | Augmentation | Epochs | Gather with grad | Precision | Local loss | Grad checkpointing | use batch-norm sync | DDP |
|---|---|---|---|---|---|---|---|---|---|---|---|---|
| R1 | 2000 | 5.0e-4 | cosine | 0.2 | RandomResizedCrop(0.8, 1.0) | 40 | ✓ | amp_bfloat16 | ✓ | ✓ | ✓ | ✗ |
| R2 | 2000 | 5.0e-4 | cosine | 0.2 | RandomResizedCrop(0.8, 1.0) | 50 | ✓ | amp_bfloat16 | ✓ | ✓ | ✓ | ✗ |
| R3 | 2000 | 5.0e-4 | cosine | 0.2 | RandomResizedCrop(0.8, 1.0) | 50 | ✓ | amp | ✗ | ✗ | ✗ | ✗ |
| R4 | 2000 | 5.0e-4 | cosine | 0.2 | RandomResizedCrop(0.8, 1.0) | 50 | ✓ | amp | ✓ | ✗ | ✗ | ✗ |
| R5 | 2000 | 5.0e-4 | Cosine | 0.2 | RandomResizedCrop(0.8, 1.0) | 50 | ✗ | amp | ✗ | ✗ | ✗ | ✗ |

Further experimentation was conducted for grad checkpointing, batch normalization synchronization between all GPUs, and enabling/disabling static graph optimization for DDP in PyTorch. It was found that these three configurations do not cause variability in the results of subsequent experiments with other similar configurations. Consequently, the use of grad checkpointing, DDP static graph, batch normalization synchronization and feature gathering was maintained throughout the rest of the experiments.

After selecting the optimized values for (a) feature gathering, (b) floating-point precision, and (c) local loss calculation, various augmentation settings for training the large transformer model were explored in "Set 2" of experiments. From here on, we make the R3 model the baseline over which further modifications are made. The implementations detail for this set are present in Table 3.

Table 3 (Set 2) Exploration of different augmentation techniques in for model training.

| Hyperparameters / Experiments | Warmup (steps) | Peak learning rate | Scheduler | Weight decay | Augmentation | Epochs |
|---|---|---|---|---|---|---|
| R6 | 2000 | 5.0e-4 | cosine | 0.2 | RandomResizedCrop(0.9, 1.0) | 50 |
| R7 | 2000 | 5.0e-4 | cosine | 0.2 | RandomResizedCrop(0.9, 1.0) | 40 |
| R8 | 2000 | 5.0e-4 | cosine | 0.2 | RandomResizedCrop(0.7, 1.0) | 50 |
| R9 | 2000 | 5.0e-4 | cosine | 0.2 | RandomResizedCrop(0.7, 1.0) | 40 |
| R10 | 2000 | 5.0e-4 | cosine | 0.2 | RandomResizedCrop(0.6, 1.0) | 50 |
| R11 | 2000 | 5.0e-4 | cosine | 0.2 | RandomResizedCrop(0.6, 1.0) | 40 |

Keeping a constant value of 0.7 for random resize crop augmentation, further experimentation with the learning rate and epochs was conducted, details of which are present in Table 4. We refer to this group of experiments as "Set 3".

Table 4 (Set 3) Exploration effect of varying learning rates and epochs in the model training.

| Hyperparameters / Experiments | Warmup (steps) | Peak learning rate | Scheduler | Weight decay | Augmentation | Epochs |
|---|---|---|---|---|---|---|
| R12 | 2000 | 1.0e-4 | cosine | 0.2 | RandomResizedCrop(0.7, 1.0) | 50 |
| R13 | 2000 | 1.0e-4 | cosine | 0.2 | RandomResizedCrop(0.7, 1.0) | 40 |
| R14 | 2000 | 5.0e-5 | cosine | 0.2 | RandomResizedCrop(0.7, 1.0) | 50 |
| R15 | 2000 | 5.0e-5 | cosine | 0.2 | RandomResizedCrop(0.7, 1.0) | 40 |
| R16 | 2000 | 1.0e-5 | cosine | 0.2 | RandomResizedCrop(0.7, 1.0) | 50 |
| R17 | 2000 | 1.0e-5 | cosine | 0.2 | RandomResizedCrop(0.7, 1.0) | 40 |

In "Set 4", as represented in Table 5, we experiment more in the lower learning rate domain with a variation in the learning rate scheduler, warmup steps and weight decay, following similar settings from [65].

*Table 5 (Set 4) Experimentation with learning rate scheduler and warmup steps for model training.*

| Hyperparameters / Experiments | Warmup (steps) | Peak learning rate | Scheduler | Weight decay | Augmentation | Epochs |
|---|---|---|---|---|---|---|
| R18 | 250 | 5.0e-5 | cosine | 0.2 | RandomResizedCrop(0.7, 1.0) | 15 |
| R19 | 250 | 5.0e-5 | constant | 0.2 | RandomResizedCrop(0.7, 1.0) | 50 |
| R20 | 250 | 5.0e-5 | constant | 0.1 | RandomResizedCrop(0.7, 1.0) | 50 |
| R21 | 250 | 1.0e-5 | cosine | 0.2 | RandomResizedCrop(0.7, 1.0) | 15 |
| R22 | 250 | 1.0e-5 | constant | 0.2 | RandomResizedCrop(0.7, 1.0) | 50 |
| R23 | 250 | 1.0e-5 | constant | 0.1 | RandomResizedCrop(0.7, 1.0) | 50 |

### 3. 6 Evaluation protocols

### 3. 6. 1 Downstream histopathology dataset

We evaluate our model on three different downstream datasets: PatchCamelyon[92] contains histopathology scans of lymph node sections labeled for metastatic tissue presence or absence as a binary label. The dataset contains 327,680 color images (96x96px) in total from which we employed the test split containing 32,678 pairs for evaluation. LC25000 (Lung) and LC25000 (Colon)[93], contain tissue of lung and adenocarcinomas, with 15,000 and 10,000 color images of size 768x768px, respectively.

### 3. 6. 2 Zero-shot Evaluation

We evaluate our model's zero-shot performance on the three datasets in Section 3.7.1. Since the model has not been trained for image classification or to take input class labels, we input actual sentences including the class names into the model, as demonstrated in Figure 3. We follow [65] while designing the input prompt templates. The prompts used for these evaluations are presented in Table 6. In this case, for each class label, we get four different input captions, which are then merged into one embedding for that particular class. Using this approach also allows for inputting 3 or more classes for classification, unlike a model trained on a fixed set of labels. In the end, the image is classified according to the sentence whose pairwise cosine similarity is the largest. This is the procedure used to generate a zero-shot linear classifier.

Table 6 For each dataset in the zero-shot image classification task, we designated classes and utilized consistent prompt templates for the classification process. The templates employed were: ["a histopathology slide showing [1]", "histopathology image of [1]", "pathology tissue showing [2]", and "presence of [2] tissue on image"], where [2] represents the class label.

| Dataset | Class Labels |
|---|---|
| PatchCamelyon | "Lymph node", "Lymph node containing metastatic tumor tissue" |
| LC 25000 (Lung) | "Benign lung", "Lung adenocarcinoma", "Lung squamous cell carcinoma" |
| LC 25000 (Colon) | "Colon adenocarcinoma", "Benign colonic tissue" |

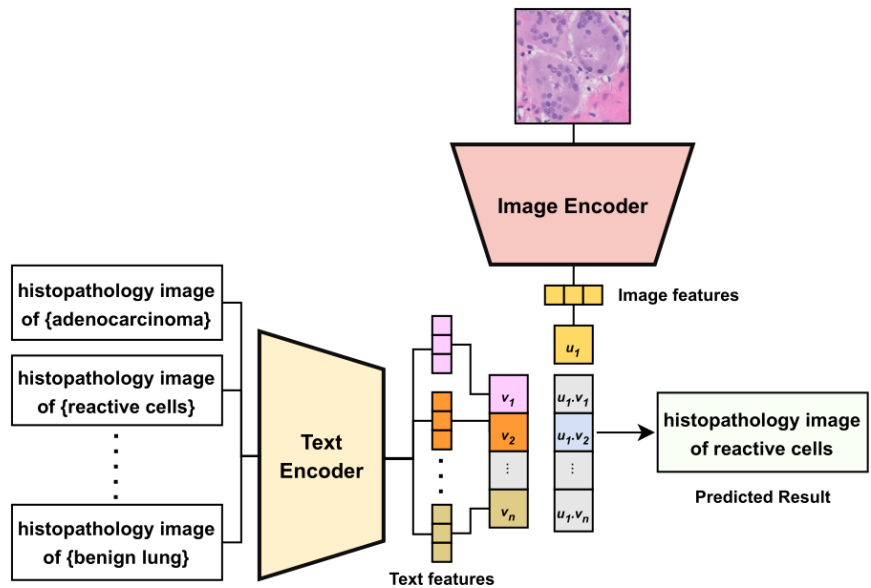

Figure 3 During the inference process, the text encoder generates a zero-shot linear classifier by transforming the names or descriptions of the classes in the target dataset into embeddings.

## 4 Experimental Results

We conducted four sets of experiments to systematically evaluate hyperparameter impacts on CLIP model performance across three histopathology datasets: LC25000 (Lung), LC25000 (Colon), and PatchCamelyon. We report accuracy for all datasets, with additional sensitivity and specificity metrics for PatchCamelyon's binary classification task.

Table 7 presents results for experiments R1-R5 (set 1), exploring fundamental configurations including precision settings and epoch variations. LC25000 (Lung) accuracies ranged from 34.92% (R4) to 75.79% (R1), while LC25000 (Colon) achieved 71.05-83.71% across experiments. PatchCamelyon showed more modest performance (46.75-59.71%) with notable sensitivity-specificity trade-offs: R4 exhibited high sensitivity (0.934) but low specificity (0.26), while R1 showed the opposite pattern. Figure 4a visualizes these performance variations across experiments and datasets.

Table 8 shows results for experiments R6-R11 (set 2), varying RandomResizedCrop parameters from 0.6 to 0.9. The 0.7 crop ratio (R8, R9) generally performed well, with R8 achieving 76.05% on LC25000 (Lung) and R9 reaching 88.67% on LC25000 (Colon). R10 (0.6 crop ratio) achieved the highest colon accuracy (91.08%) but lowest lung accuracy (56.28%), demonstrating dataset-specific augmentation responses. PatchCamelyon accuracies ranged from 50.29-59.80% with varying sensitivity-specificity patterns. Figure 4b illustrates these augmentation effects.

Table 9 presents experiment R12-R17 (set 3) exploring learning rates from 1.0e-5 to 1.0e-4. Very low learning rates (R12, R14, R15) resulted in poor performance and extremely low sensitivity (≤0.015) on PatchCamelyon, indicating underfitting. Figure 4c demonstrates the critical impact of learning rate selection on model performance.

Table 10 shows experiments R18-R23 (set 4) investigating learning rate schedulers and weight decay with reduced warmup steps. Performance was relatively stable across configurations, with LC25000 (Lung) accuracies ranging 66.36-73.45% and LC25000 (Colon) reaching up to 88.12% (R19). Most PatchCamelyon experiments showed high specificity (>0.93) but low sensitivity (<0.25), with R19 providing the most balanced trade-off. Figure 4d visualizes these scheduler and weight decay effects.

Table 7 Experimental results for "Set 1" (Base model configurations)

| Experiment | LC25000 (Lung) | LC25000 (Colon) | PatchCam | | |
|---|---|---|---|---|---|
| | Accuracy (%) | Accuracy (%) | Accuracy (%) | Sensitivity | Specificity |
| R1 | 75.79 | 71.05 | 48.18 | 0.096 | 0.867 |
| R2 | 66.84 | 78.69 | 51.65 | 0.257 | 0.776 |
| R3 | 75.35 | 83.64 | 47.47 | 0.48 | 0.47 |
| R4 | 34.92 | 73.80 | 59.71 | 0.934 | 0.26 |
| R5 | 67.41 | 83.71 | 46.75 | 0.572 | 0.363 |

Table 8 Experimental results for "Set 2" (Augmentation variations)

| Experiment | LC25000 (Lung) | LC25000 (Colon) | PatchCam | | |
|---|---|---|---|---|---|
| | Accuracy (%) | Accuracy (%) | Accuracy (%) | Sensitivity | Specificity |
| R6 | 63.36 | 60.34 | 50.29 | 0.826 | 0.180 |
| R7 | 57.72 | 75.65 | 53.58 | 0.603 | 0.469 |
| R8 | 76.05 | 88.44 | 57.72 | 0.913 | 0.242 |
| R9 | 69.93 | 88.67 | 56.48 | 0.719 | 0.411 |
| R10 | 56.28 | 91.08 | 59.80 | 0.327 | 0.871 |
| R11 | 57.16 | 87.81 | 51.20 | 0.657 | 0.367 |

Table 9 Experimental results for "Set 3" (Learning rate variations)

| Experiment | LC25000 (Lung) | LC25000 (Colon) | PatchCam | | |
|---|---|---|---|---|---|
| | Accuracy (%) | Accuracy (%) | Accuracy (%) | Sensitivity | Specificity |
| R12 | 59.51 | 72.95 | 49.61 | 0.004 | 0.988 |
| R13 | 77.95 | 89.79 | 47.58 | 0.201 | 0.751 |
| R14 | 68.86 | 70 | 50.31 | 0.015 | 0.991 |
| R15 | 48.1 | 73.31 | 49.87 | 0.001 | 0.996 |
| R16 | 58.69 | 69.63 | 56.69 | 0.268 | 0.864 |
| R17 | 59.22 | 73.95 | 57.46 | 0.333 | 0.816 |

Table 10 Experimental results for "Set 4" (Effect of Learning rate scheduler, Weight decay)

| Experiment | LC25000 (Lung) | LC25000 (Colon) | PatchCam | | |
|---|---|---|---|---|---|
| | Accuracy (%) | Accuracy (%) | Accuracy (%) | Sensitivity | Specificity |
| R18 | 73.45 | 72.72 | 54.15 | 0.091 | 0.991 |
| R19 | 51.80 | 88.12 | 61.75 | 0.438 | 0.797 |
| R20 | 70.19 | 67.80 | 56.79 | 0.243 | 0.859 |
| R21 | 66.73 | 59.05 | 56.79 | 0.205 | 0.931 |
| R22 | 66.36 | 70.30 | 52.77 | 0.070 | 0.985 |
| R23 | 66.41 | 68.61 | 52.98 | 0.074 | 0.985 |

Figure 4 visualizes the obtained results for all our models, leading to insights about each configuration.

(a)

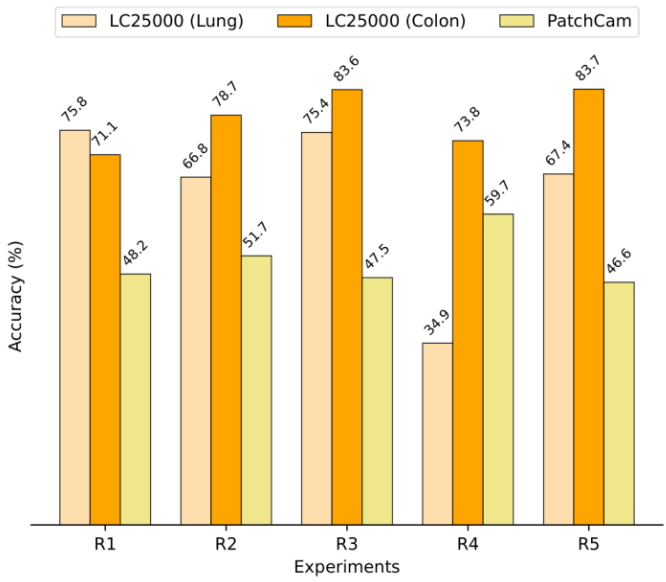
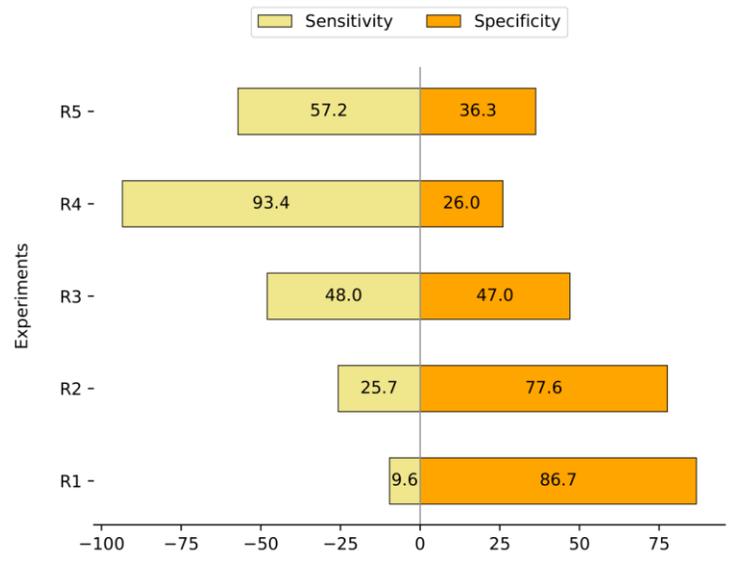

(b)

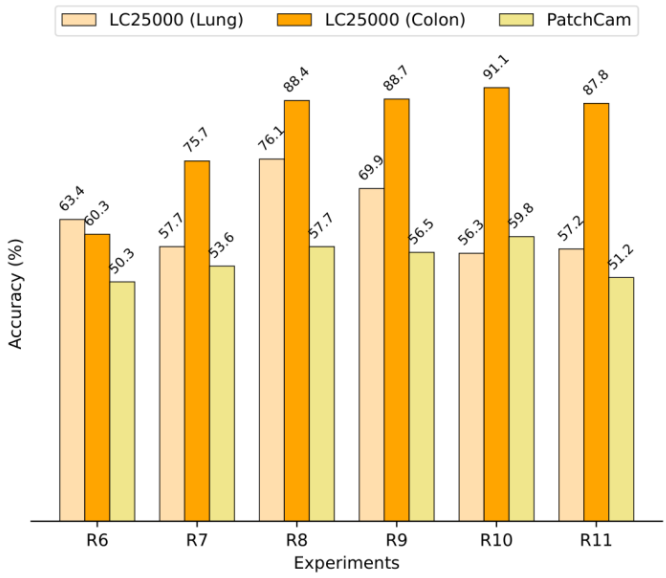
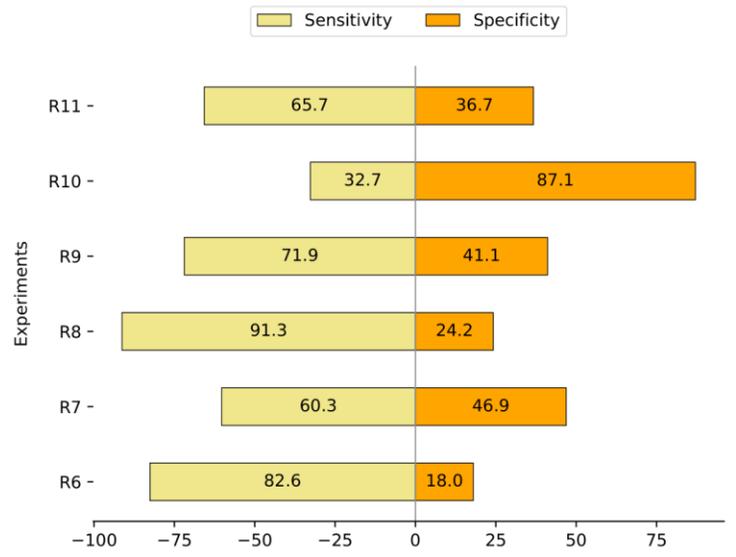

(c)

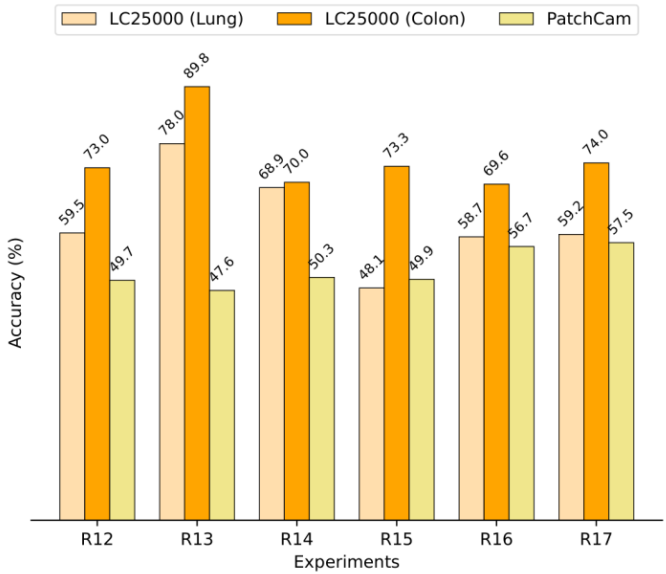
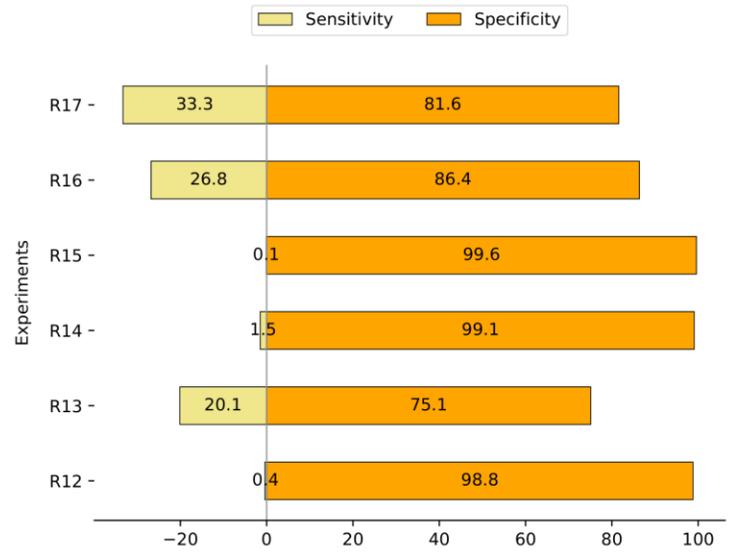

(d)

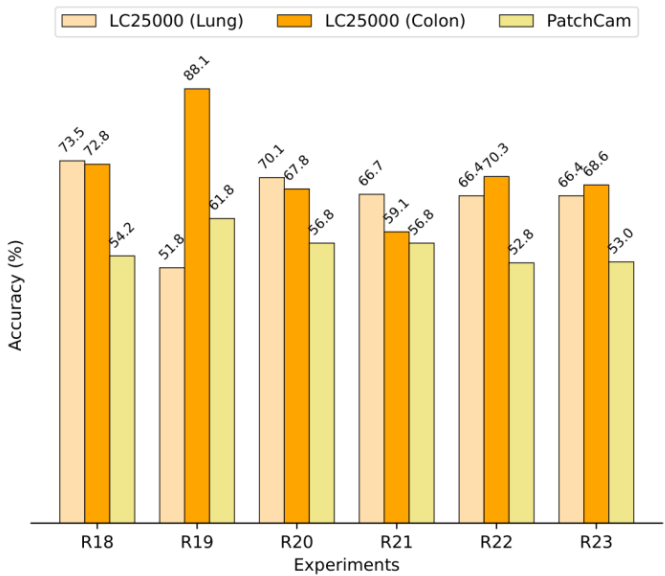
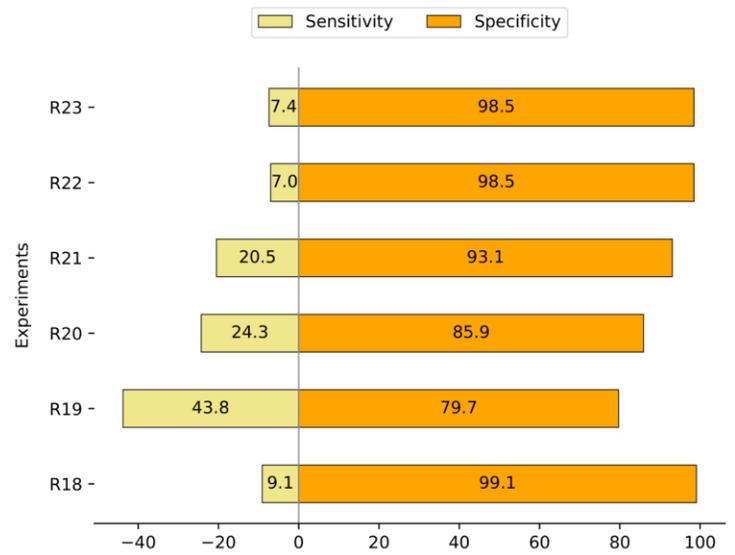

Figure 4: (a), (b), (c) and (d) represent the result of Set 1, 2, 3 and 4, respectively.

## 5 Discussion

This study finds its place between the efforts to ensure reproducible deep learning models and the computational histopathology domain. Despite encountering significant variability in our results, which poses challenges for reproducibility, we have identified several consistent lessons that inform best practices for future work in the field. Below, we organize the findings into key insights derived from each experiment set.

From Set 1, we observed that the R4 configuration, which combined local loss with feature gathering, consistently underperformed across all datasets. In contrast, gathering features from all nodes without local loss (R3) yielded promising results, suggesting that local loss introduces noise or instability in distributed training setups. This finding highlights the importance of carefully evaluating distributed training strategies, as seemingly minor design choices can compromise reproducibility and reliability.

In Set 2, we found that random resized crop values of 0.9 and 0.6 produced inferior results compared to 0.7 and 0.8. Notably, the 0.7 crop ratio (R9) yielded balanced and consistent performance across datasets, especially for the LC25000 (Colon) dataset, which peaked at 88.7% accuracy (R9). These results demonstrate that augmentation hyperparameters are pivotal factors influencing model generalization and reproducibility. For histopathology, where morphological patterns can be subtle, overly aggressive or insufficient cropping may distort key features.

Across Sets 3 and 4, experiments with lower learning rates consistently degraded performance on all datasets, with pronounced skewness in sensitivity and specificity for PatchCamelyon. Moreover, neither alternative schedulers nor adjustments in warmup steps and weight decay mitigated this decline, reinforcing the conclusion that certain learning rate ranges might be unsuitable for reproducible training in this domain.

Despite overall variability, the LC25000 (Colon) dataset consistently exhibited higher and more stable accuracies across multiple experiment sets. This indicates that certain dataset characteristics (e.g., sample diversity, label quality, pathology subtype homogeneity) can influence reproducibility outcomes. Identifying "stable" benchmark datasets could therefore help anchor reproducibility studies and serve as a reference point for comparing configurations.

Overall, these lessons highlight that reproducibility in deep learning for histopathology depends on more than fixing seeds and documenting hyperparameters. Certain configurations—such as augmentation choices, local vs. global loss settings, and learning rate ranges—directly determine whether results are stable and generalizable. This means that researchers in this domain should also put effort into systematically evaluating hyperparameter domains that either stabilize or destabilize results, in addition to being transparent about the model configurations.

## 6 Limitations

Although we believe that our study highlights the issue of reproducing results in deep learning and demonstrates how to optimize best and report experimental results and model implementations to ensure the experiments conducted were limited to a specific set of configurations and datasets, which can be expanded upon. This constrains the generalizability of our findings across diverse types of image classification tasks and datasets. Moreover, for completeness in terms of our reported results, the accuracy and trade-offs observed may be precise to the datasets used, limiting the applicability of conclusions drawn to other, more specific datasets or real-world conditions.

## 7 Conclusion

In conclusion, this study highlights the pivotal role of hyperparameter selection and data augmentation in the training of foundation models for histopathology within deep learning. By training a CLIP model on the Quilt-1M dataset and exploring various configurations for the CLIP model, we not only address the reproducibility challenges posed by software randomness and hardware non-determinism but also establish a methodical framework for hyperparameter documentation. This framework is crucial for facilitating reproducibility in future research, highlighting the importance of detailed reporting and strategic tuning in developing reliable AI models for clinical applications.